\documentclass[letterpaper]{article} 
\usepackage{aaai23}  
\usepackage{times}  
\usepackage{helvet}  
\usepackage{courier}  
\usepackage[hyphens]{url}  
\usepackage{graphicx} 
\usepackage{natbib}  
\usepackage{caption} 
\usepackage{algorithm}
\usepackage{algorithmic}

\usepackage{amsmath,amssymb} 
\usepackage{adjustbox}
\usepackage{multirow}
\usepackage{booktabs}
\usepackage{subfigure}
\usepackage{pifont}
\newcommand{\xmark}{\ding{55}}
\usepackage{color,colortbl}

\def\ul#1{\underline{#1}}

\usepackage{lipsum}

\newcommand\blfootnote[1]{%
\begingroup
\renewcommand\thefootnote{}\footnote{#1}%
\addtocounter{footnote}{-1}%
\endgroup
}

\usepackage{newfloat}
\usepackage{listings}
\DeclareCaptionStyle{ruled}{labelfont=normalfont,labelsep=colon,strut=off} 
\floatstyle{ruled}
\newfloat{listing}{tb}{lst}{}
\floatname{listing}{Listing}
\title{Few-Shot Object Detection via Variational Feature Aggregation}
\author{
    Jiaming Han\textsuperscript{\rm 1,2}\thanks{Work done during internship at Tencent YouTu Lab.}, Yuqiang Ren\textsuperscript{\rm 3}, Jian Ding\textsuperscript{\rm 1,2}, Ke Yan\textsuperscript{\rm 3$\dag$}, Gui-Song Xia\textsuperscript{\rm 1,2$\dag$}\blfootnote{Corresponding author.}
}
\affiliations{
    \textsuperscript{\rm 1}NERCMS, School of Computer Science, Wuhan University\\ 
    \textsuperscript{\rm 2}State Key Lab. LIESMARS, Wuhan University\\
    \textsuperscript{\rm 3}YouTu Lab, Tencent
    
    \{hanjiaming, jian.ding, guisong.xia\}@whu.edu.cn, 
    \{condiren, kerwinyan\}@tencent.com
}

\usepackage{bibentry}

\begin{document}

\maketitle

\begin{abstract}
As few-shot object detectors are often trained with abundant base samples and fine-tuned on few-shot novel examples,
the learned models are usually biased to base classes and sensitive to the variance of novel examples.
To address this issue, we propose a meta-learning framework with two novel feature aggregation schemes.
More precisely, we first present a Class-Agnostic Aggregation (CAA) method, where the query and support features can be aggregated regardless of their categories.
The interactions between different classes encourage class-agnostic representations and reduce confusion between base and novel classes.
Based on the CAA, we then propose a Variational Feature Aggregation (VFA) method, which encodes support examples into class-level support features for robust feature aggregation. 
We use a variational autoencoder to estimate class distributions and sample variational features from distributions that are more robust to the variance of support examples.
Besides, we decouple classification and regression tasks so that VFA is performed on the classification branch without affecting object localization.
Extensive experiments on PASCAL VOC and COCO demonstrate that our method significantly outperforms a strong baseline (up to 16\%) and previous state-of-the-art methods (4\% in average). Code will be available at: \url{https://github.com/csuhan/VFA}
\end{abstract}

\section{Introduction}
\label{sec:intro}

This paper studies the problem of few-shot object detection (FSOD), a recently-emerged challenging task in computer vision~\cite{yan2019meta,kang2019few}. Different from generic object detection~\cite{girshick2014rich,redmon2016you,ren2017faster}, FSOD assumes that we have abundant samples of some base classes but only a few examples of novel classes. Thus, a dynamic topic is how to improve the recognition capability of FSOD on novel classes by transferring the knowledge of base classes to novel ones.

\begin{figure}
    \centering
    \includegraphics[width=\linewidth]{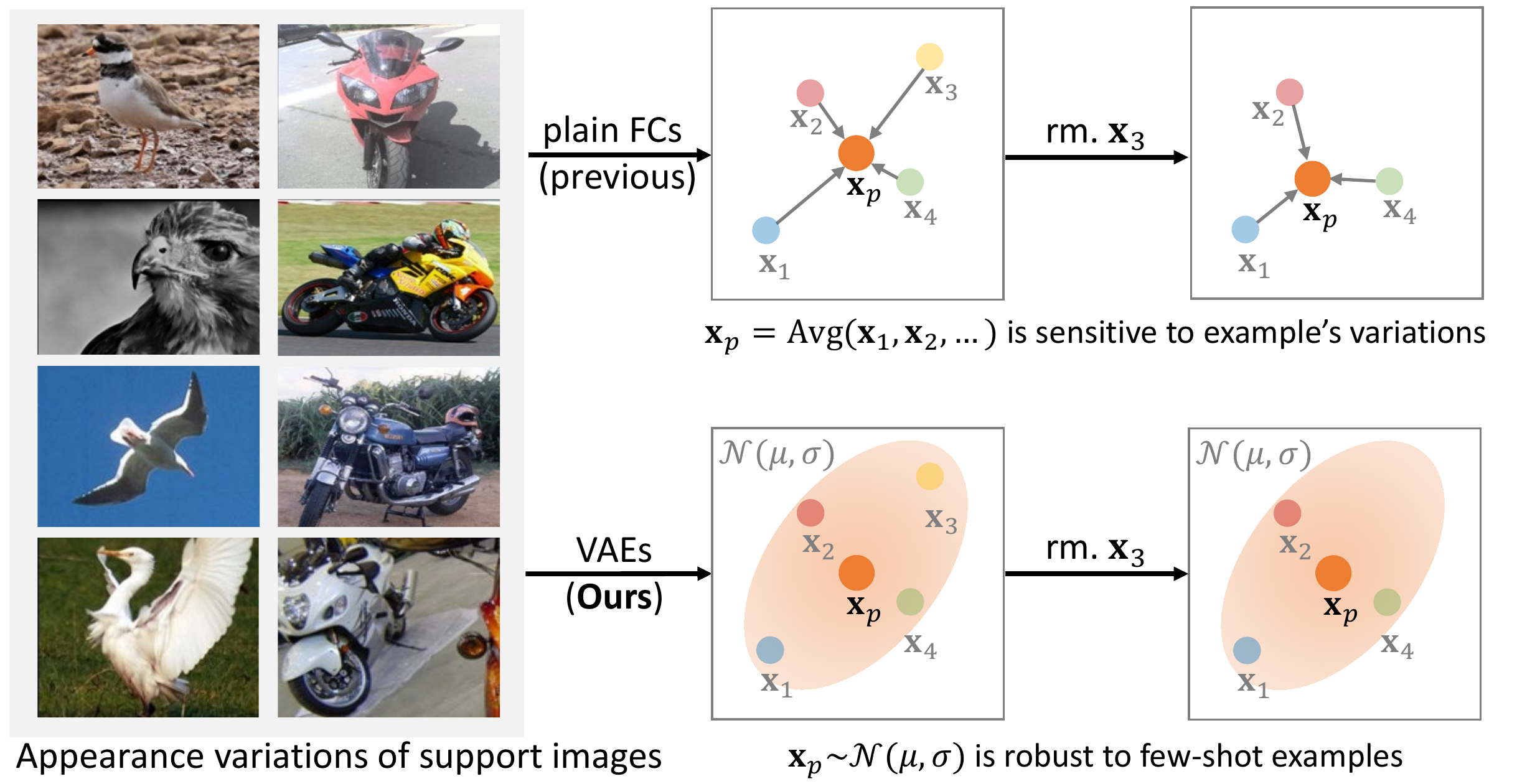}
    \caption{Comparisons of different support feature encoding methods. Previous methods use plain fully-connected (FC) layers to encode support features and obtain class prototypes by averaging these features: $\mathbf{x}_p={\rm Avg}(\mathbf{x}_1,\mathbf{x}_2,\dots)$.
    In contrast, our method uses variational autoencoders (VAEs) pre-trained on abundant base examples to estimate the distributions of novel classes. 
    Since intra-class variance is shared across classes and can be modeled with common distributions~\cite{lin2018deep}, we use a shared VAE to transfer the distributions of base classes to novel classes.
    Finally, we can sample class prototypes $\mathbf{x}_p$ from the distributions $\mathcal{N}(\mu, \sigma)$ that are robust to the variance of few-shot examples. rm.: remove.}
    \label{fig:vfa_illustration}
\end{figure}

In general, FSOD follows a two-stage training paradigm. In stage-I, the detector is trained with abundant base samples to learn generic representations required for the object detection task, such as object localization and classification. 
In stage-II, the detector is fine-tuned with only $K$ shots ($K$=1, 2, 3, $\dots$) novel examples. 
Despite the great success of this paradigm, the learned models are usually biased to base classes due to the imbalance between base and novel classes.
As a result, the model will confuse novel objects with similar base classes.
See Fig.~\ref{fig:confusion_matrix} (top) for an instance, the novel class, {\em cow}, has high similarities with several base classes such as {\em dog, horse} and {\em sheep}.
Besides, the model is sensitive to the variance of novel examples. Since we only have $K$ shots examples per class, the performance highly depends on the quality of the support sets.
As shown in Fig.~\ref{fig:vfa_illustration}, appearance variations are common in FSOD. 
Previous methods~\cite{yan2019meta} consider each support example as a single point in the feature space and average all features as class prototypes.
However, it is difficult to estimate the real class centers with a few examples.

In this paper, we propose a meta-learning framework to address this issue. 
Firstly, we build a strong meta-learning baseline based on Meta R-CNN~\cite{yan2019meta}, which even outperforms a representative two-stage fine-tuning approach TFA~\cite{wang2020frustratingly}. 
By revisiting the feature aggregation module in meta-learning frameworks, we propose Class-Agnostic Aggregation (CAA) and Variational Feature Aggregation (VFA) to reduce class bias and improve the robustness to example's variances, respectively.

Feature aggregation is a crucial design in FSOD, which defines how query and support examples interact. Previous works such as Meta R-CNN adopt a class-specific aggregation scheme (Fig.~\ref{fig:feature_aggregation} (a)), \emph{i.e.}, query features are aggregated with support features of the same class, ignoring cross-class interactions.
In contrast, we propose CAA (Fig.~\ref{fig:feature_aggregation} (b)) which allows feature aggregation between different classes.
Since CAA encourages the model to learn class-agnostic representations, the bias towards base classes is reduced.
Besides, the interactions between different classes simultaneously model class relations so that novel classes will not be confused with base classes.

Based on CAA, we propose VFA which encodes support examples into class-level support features. 
Our motivation is that intra-class variance (\emph{e.g.} appearance variations) is shared across classes and can be modeled with common distributions~\cite{lin2018deep}.
So we can use base classes' distributions to estimate novel classes' distributions.
We achieve this by modeling each class as a common distribution with variational autoencoders (VAEs).
We firstly train the VAE on abundant base examples and then fine-tune it on few-shot novel examples.
By transferring the learned intra-class variance to novel classes, our method can estimate novel classes' distributions with only a few examples (Fig.~\ref{fig:vfa_illustration}).
Finally, we sample support features from distributions and aggregate them with query features to produce more robust predictions.

We also propose to decouple classification and regression tasks so that our feature aggregation module can focus on learning translation-invariant features without affecting object localization.
We conduct extensive experiments on two FSOD datasets, PASCAL VOC~\cite{everingham2010pascal} and COCO~\cite{lin2014microsoft} to demonstrate the effectiveness of our method. We summarize our contributions as follows:
\begin{itemize}
    \item We build a strong meta-learning baseline Meta R-CNN++ and propose a simple yet effective Class-Agnostic Aggregation (CAA) method.
    \item We propose Variational Feature Aggregation (VFA), which transforms instance-wise features into class-level features for robust feature aggregation. To our best knowledge, we are the first to introduce variational feature learning into FSOD.
    \item Our method significantly improves the baseline Meta R-CNN++ and achieves a new state-of-the-art for FSOD. For example, we outperform the strong baseline by 9\%$\sim$16\% and previous best results by 3\%$\sim$7\% on the Novel Set 1 of PASCAL VOC.
\end{itemize}

\section{Related Work}

\noindent {\bf Generic Object Detection.} Object detection has witnessed significant progress in the past decade,
which can be roughly divided into two groups: one-stage and two-stage detectors. 
One-stage detectors predict bounding boxes and class labels by presetting dense anchor boxes~\cite{redmon2016you,liu2016ssd,lin2017focal}, points~\cite{law2018cornernet,zhou2019objects}, or directly output sparse predictions~\cite{carion2020end,chen2021pix2seq}. Two-stage detectors~\cite{girshick2014rich,girshick2015fast,ren2017faster} first generate a set of object proposals with Region Proposal Network (RPN) and then perform proposal-wise classification and regression. However, most generic detectors are trained with abundant samples and not designed for data-scarce scenarios.


\noindent {\bf Few-Shot Object Detection.}
Early attempts~\cite{kang2019few,yan2019meta,wang2019meta} in FSOD adopt \textbf{\textit{meta-learning}} architectures. 
FSRW~\cite{kang2019few} and Meta R-CNN~\cite{yan2019meta} aggregate image/RoI-level query features with support features generated by a meta learner. 
Following works explore different designs of meta-learning architectures, \emph{e.g.}, feature aggregation scheme~\cite{xiao2020few,fan2020few,hu2021dense,zhang2021accurate,han2021query} and feature space augmentation~\cite{li2021beyond,li2021transformation}. 
Different from meta-learning, Wang et al. propose a simple two-stage \textbf{\textit{fine-tuning}} approach, TFA~\cite{wang2020frustratingly}.
TFA shows that only fine-tuning the last layers can significantly improve the FSOD performance. Due to the simple structure of TFA, a line of works~\cite{sun2021fsce,zhu2021semantic,qiao2021defrcn,cao2021few} following TFA are proposed.
\textbf{\textit{In this work}}, we build a strong meta-learning baseline that even surpasses the fine-tuning baseline TFA. Then we revisit the feature aggregation scheme and propose two novel feature aggregation methods, CAA and VFA, achieving a new state-of-the-art in FSOD.

\noindent {\bf Variational Feature Learning.}
Given an input image/feature, we can transform it into a distribution with VAEs. By sampling features from the distribution, we can model intra-class variance that defines the class's character. The variational feature learning paradigm has been used in various tasks, \emph{e.g.}, zero/few-shot learning~\cite{zhang2019variational,xu2021variational,kim2019variational}, metric learning~\cite{lin2018deep} and disentanglement learning~\cite{ding2020guided}.
In this work, we use VAEs trained on abundant base examples to estimate novel classes' distributions with only a few examples.
Besides, we also propose a consistency loss to make the model produce class-specific distributions. To our best knowledge, we are the first to introduce variational feature learning into FSOD.

\section{Background and Meta R-CNN++}

\subsection{Preliminaries}

\noindent {\bf Problem Definition.} We follow the FSOD settings in previous works~\cite{yan2019meta,wang2020frustratingly}. 
Assume we have a dataset $D=\{(x,y), x \in X, y \in Y\}$ with a set of classes $C$, where $x$ is the input image and $y=\{c_i, \mathbf{b}_i\}_{i=1}^N$ is the corresponding class label $c$ and bounding box $\mathbf{b}$ annotations. 
We then split the dataset into base classes $C_b$ and novel classes $C_n$ where $C_b \cup C_n = C$ and $C_b \cap C_n = \varnothing$. 
Generally, we have abundant samples of $C_b$ and $K$ shots samples of $C_n$ ($K$=1, 2, 3, \dots). The goal is to detect objects of $C_n$ with only $K$ shots annotated instances. 
Existing few-shot detectors usually adopt a two-stage training paradigm: base training and few-shot fine-tuning, where the representations learned from $C_b$ are transferred to detect novel objects in the fine-tuning stage.

\noindent {\bf Meta-Learning Based FSOD.} We take Meta R-CNN~\cite{yan2019meta} for an example. As shown in Fig.~\ref{fig:vfa}, the main framework is a siamese network with a query feature encoder $\mathcal{F}_Q$, a support feature encoder $\mathcal{F}_S$, a feature aggregator $\mathcal{A}$ and a detection head $\mathcal{F}_D$. 
Typically, $\mathcal{F}_Q$ and $\mathcal{F}_S$ share most parameters and $\mathcal{A}$ refers to the channel-wise product operation.
Meta R-CNN follows the episodic training paradigm~\cite{vinyals2016matching}. Each episode is composed of a set of support images and binary masks of annotated objects, $\{x_i, M_i\}_{i=1}^N$, where $N$ is the number of training classes.
Specifically, we first feed the support set $\{x_i, M_i\}_{i=1}^N$ to $\mathcal{F}_S$ to generate class-specific support features $\{S_i\}_{i\in C}$, and the query image to $\mathcal{F}_Q$ to generate a set of RoI features $\{Q^m\}$ ($m$ is the index of RoIs).
Then we aggregate each $Q^m$ and $S_i$ with the feature aggregator $\mathcal{A}$.
Finally, the aggregated features $\widetilde{Q}_i^m$ are fed to the detection head $\mathcal{F}_D$ to produce final predictions.

\subsection{Meta R-CNN++: Stronger Meta-Learning Baseline}
\label{subsec:meta_rcnn_plus}

\begin{table}[t]
\begin{center}
\small
\begin{tabular}{@{}lcccccc@{}}
\toprule
setting                 & TFA          &Meta R-CNN$^*$& \multicolumn{3}{c}{Meta R-CNN++}            \\ \midrule
param freeze            & \checkmark   & \xmark     & \checkmark   & \checkmark   & \checkmark    \\
cosine cls.       & \checkmark   & \xmark     & \xmark       & \checkmark   & \checkmark    \\ 
last layer init.        & copy         & rand       & rand         & rand         & copy          \\ \midrule
bAP (stage-I)           & {\bf 80.8}   & 72.8       & 77.6         & 77.6         & 77.6          \\
bAP (stage-II)          & 79.6         & 47.4       & 64.9         & 68.2         & {\bf 76.8}    \\ 
nAP                     & 39.8         & 20.7       & {\bf 42.0}   & 40.5         & 41.6          \\
\bottomrule
\end{tabular}
\end{center}
    \caption{Difference analysis between Meta R-CNN and TFA.
    The results are evaluated under the 1 shot setting of PASCAL VOC Novel Set 1. stage-I and stage-II: base training and fine-tuning stages. $^*$: Our re-implemented results.}
    \label{tab:meta_rcn_plus}
    \end{table}

Meta-learning has proved a promising approach, but the fine-tuning based approach receives more and more attention recently due to its superior performance. 
Here we aim to bridge the gap between the two approaches. We choose Meta R-CNN and TFA as baselines and explore how to build a strong FSOD baseline with meta-learning.

Although both methods follow a two-stage training paradigm, TFA optimizes the model with advanced techniques in the fine-tuning stage:
{\bf (a)} TFA freezes most network parameters, and only trains the last classification and regression layers so that the model will not overfit to few-shot examples.
{\bf (b)} Instead of randomly initializing the classification layer, TFA copies pre-trained weights of base classes and only initializes the weights of novel classes.
{\bf (c)} TFA adopts cosine classifier~\cite{gidaris2018dynamic} rather than a linear classifier.

Considering the success of TFA, we build Meta R-CNN++, which follows the architecture of Meta R-CNN but aligns most hyper-parameters with TFA.
Here we explore different design choices to mitigate the gap between the two approaches, shown in Tab.~\ref{tab:meta_rcn_plus}.
{\bf (a) Parameter freeze.} By adopting the same parameter freezing strategy, Meta R-CNN++ significantly outperforms Meta R-CNN and even achieves higher novel AP than TFA.
{\bf (b) Cosine classifier.} Different from TFA, Meta R-CNN++ with the cosine classifier does not surpass the linear classifier in nAP (41.6 \emph{vs.} 42.0), but its performance on base classes is better than the linear classifier (68.2 \emph{vs.} 64.9). 
{\bf (c) Alleviate base forgetting.} 
We follow TFA and copy the pre-trained classifier weights of base classes. We find Meta R-CNN++ can also maintain the performance on base classes (76.8 \emph{vs.} 77.6).

The above experiments indicate that meta-learning remains a promising approach for FSOD as long as we carefully handle the fine-tuning stage. Therefore, we choose Meta R-CNN++ as our baseline in the following sections.

\section{The Proposed Approach}


\begin{figure}[t]
    \centering
    \includegraphics[width=\linewidth]{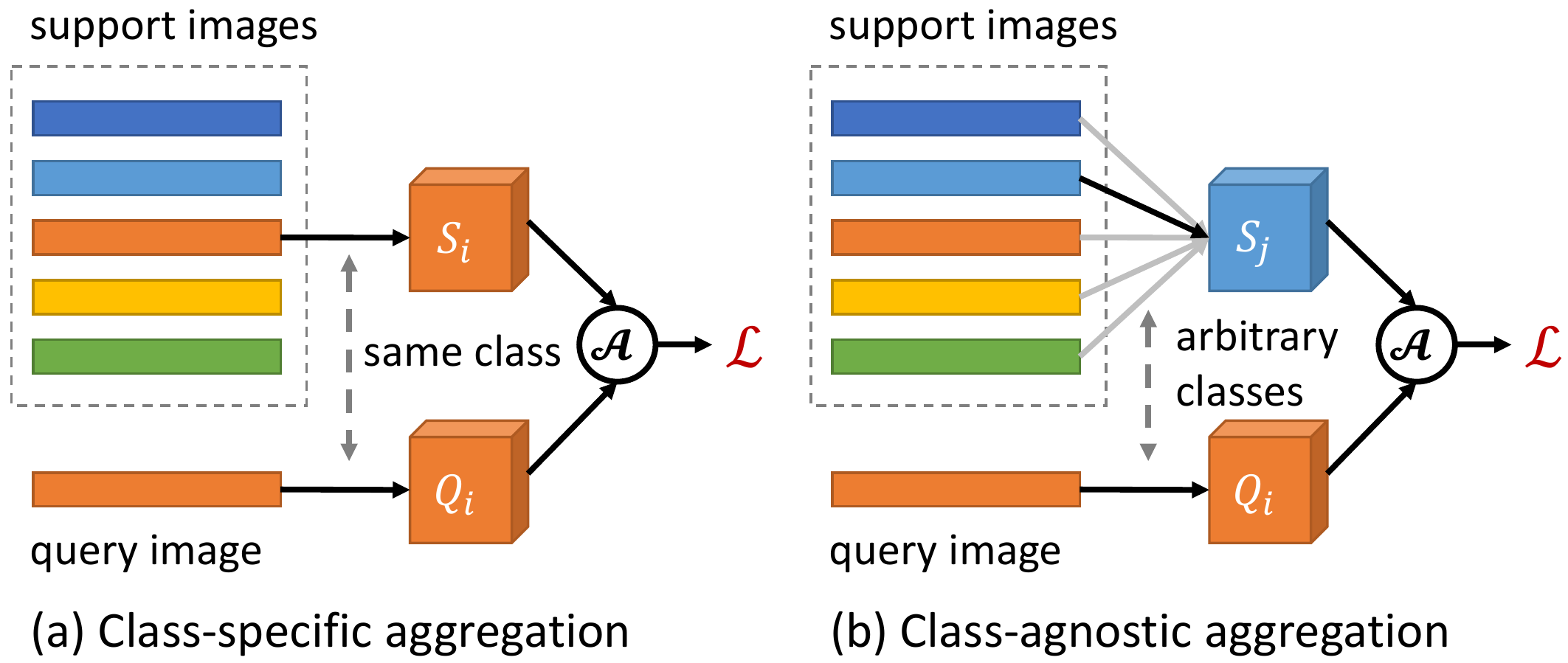}
    \caption{Illustration of two feature aggregation methods. $S_i$/$Q_i$: support and query features of class $i$. $\mathcal{A}$: feature aggregation. $\mathcal{L}$: loss functions.}
    \label{fig:feature_aggregation}
    \end{figure}

\begin{figure*}[t]
    \centering
    \includegraphics[width=0.9\textwidth]{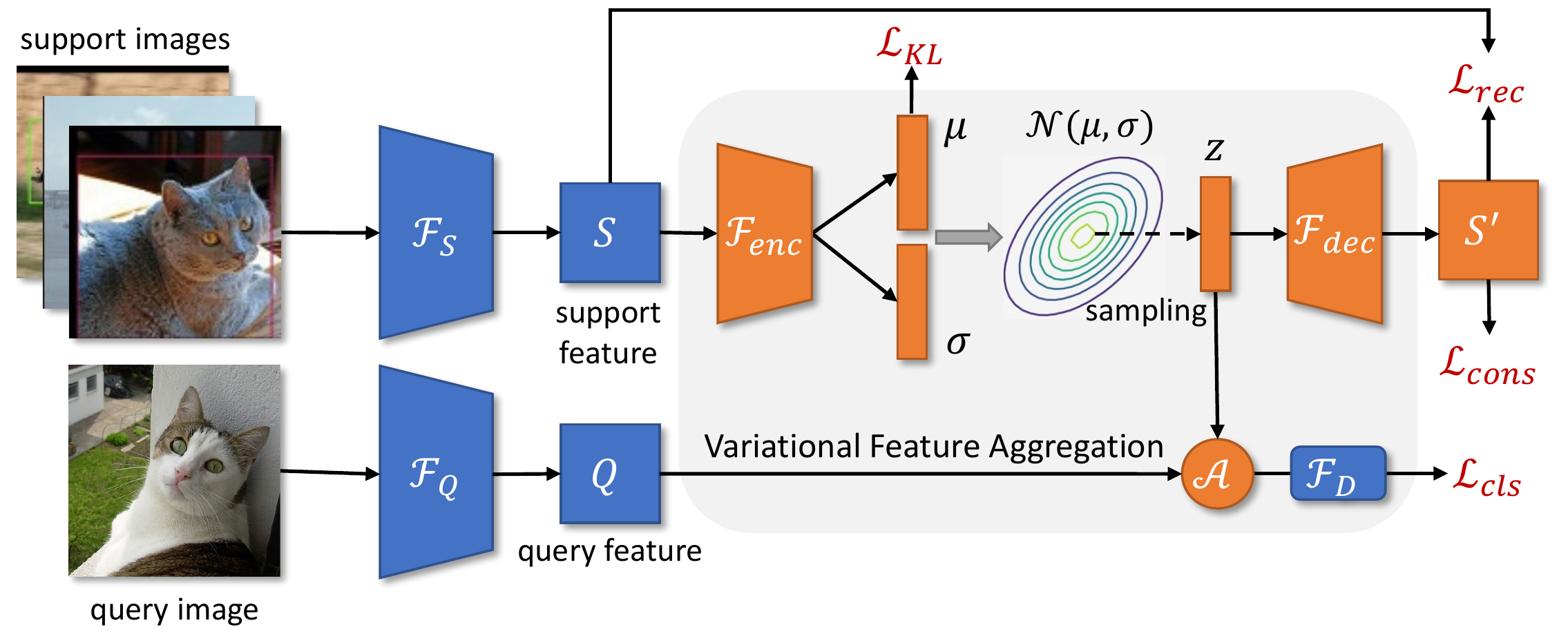}
    \caption{Overview of our framework. $\mathcal{F}_{Q}$ and $\mathcal{F}_S$ denote query and support feature extractors, respectively. $\mathcal{F}_{enc}$ and $\mathcal{F}_{dec}$ are the variational feature encoder and decoder. $\mathcal{F}_D$: the detection head. $\mathcal{A}$: feature aggregation. Note that we do not visualize RPN and the regression branch for simplicity.}
    \label{fig:vfa}
\end{figure*}

\subsection{Class-Agnostic Aggregation}
\label{subsec:caa}

Feature aggregation is an important module in meta-learning based FSOD~\cite{kang2019few,yan2019meta}. Many works adopt a class-specific aggregation (CSA) scheme. 
Let us assume that a query image has an object of class $C_Q=\{i\}$ and the corresponding RoI features $\{Q^m_i\}$. In the training phase, as shown in Fig.~\ref{fig:feature_aggregation} (a), CSA aggregates each RoI feature $Q_i^m$ with the support features $S_i$ of the same class: $\widetilde{Q}_{ii}^m = \mathcal{A}(Q_i^m, S_i)$. 
In the testing phase, CSA aggregates the RoI feature with support features of all classes: $\widetilde{Q}_{ij}^m = \mathcal{A}(Q_i^m, S_j), j\in C$, and each support feature $S_j$ is to predict objects of its corresponding class. 
Notably, if the query image contains multiple classes, CSA aggregates the query features with each support feature in $C_Q$: $\widetilde{Q}_{ij}^m = \mathcal{A}(Q_i^m, S_j), j\in C_Q$. But CSA still follows the class-specific way, as support features not belonging to $C_Q$ will never be aggregated with the query feature.

As discussed before, the learned models are usually biased to base classes due to the imbalance between base and novel classes.
Therefore, we revisit CSA and propose a simple yet effective Class-Agnostic Aggregation (CAA). 
See Fig.~\ref{fig:feature_aggregation} (b) for an instance, CAA allows feature aggregation between different classes, which encourages the model to learn class-agnostic representations and thereby reduces the class bias.
Besides, the interactions between different classes can simultaneously model class relations so that novel classes will not confuse with base classes.
Formally, for each RoI feature $Q_i^m$ of class $i \in C$ and a set of support features $\{S_j\}_{j \in C}$, we {\bf randomly} select a support feature $S_{j^*}$ of class $j^*$ to aggregate with the query feature,
\begin{align}\label{eq:caa}
    \widetilde{Q}_{ij^*}^m = \mathcal{A}(Q_i^m, S_{j^*}), j^* \in C.
\end{align}
Then we feed the aggregated feature $\widetilde{Q}_{ij^*}^m$ to the detection head $\mathcal{F}_D$ to output classification scores $\mathbf{p}=\mathcal{F}_D(\widetilde{Q}_{ij^*}^m)$, which is supervised with the label of class $i$.
Note that CAA is used for training; the testing phase still follows CSA.


\begin{table*}[t]
    \centering
\resizebox{\textwidth}{!}{
\setlength{\tabcolsep}{3.5pt}
\small
\begin{tabular}{@{}l|c|ccccc|ccccc|ccccc|c@{}}
\toprule
\multirow{2}{*}{Method / Shots}        & \multirow{2}{*}{Backbone}   & \multicolumn{5}{c|}{Novel Set 1}                                         & \multicolumn{5}{c|}{Novel Set 2}                                         & \multicolumn{5}{c|}{Novel Set 3}                                       & \multirow{2}{*}{Avg.}    \\ 
                                       &                             & 1            & 2            & 3            & 5            & 10           & 1            & 2            & 3            & 5            & 10           & 1            & 2            & 3            & 5            & 10        &    \\ \midrule
FSRW~\cite{kang2019few}                & YOLOv2                      & 14.8         & 15.5         & 26.7         & 33.9         & 47.2         & 15.7         & 15.3         & 22.7         & 30.1         & 40.5         & 21.3         & 25.6         & 28.4         & 42.8         & 45.9      & 28.4   \\
MetaDet (Wang et al. 2019)             & VGG16                       & 18.9         & 20.6         & 30.2         & 36.8         & 49.6         & 21.8         & 23.1         & 27.8         & 31.7         & 43.0         & 20.6         & 23.9         & 29.4         & 43.9         & 44.1      & 31.0   \\
Meta R-CNN~\cite{yan2019meta}          & ResNet-101                  & 19.9         & 25.5         & 35.0         & 45.7         & 51.5         & 10.4         & 19.4         & 29.6         & 34.8         & 45.4         & 14.3         & 18.2         & 27.5         & 41.2         & 48.1      & 31.1   \\
TFA w/ cos~\cite{wang2020frustratingly}& ResNet-101                  & 39.8         & 36.1         & 44.7         & 55.7         & 56.0         & 23.5         & 26.9         & 34.1         & 35.1         & 39.1         & 30.8         & 34.8         & 42.8         & 49.5         & 49.8      & 39.9   \\
MPSR~\cite{wu2020multi}                & ResNet-101                  & 41.7         & -            & 51.4         & 55.2         & 61.8         & 24.4         & -            & 39.2         & 39.9         & 47.8         & 35.6         & -            & 42.3         & 48.0         & 49.7      & -   \\
Retentive~\cite{fan2021generalized}    & ResNet-101                  & 42.4         & 45.8         & 45.9         & 53.7         & 56.1         & 21.7         & 27.8         & 35.2         & 37.0         & 40.3         & 30.2         & 37.6         & 43.0         & 49.7         & 50.1      & 41.1   \\
Halluc~\cite{zhang2021hallucination}   & ResNet-101                  & 47.0         & 44.9         & 46.5         & 54.7         & 54.7         & 26.3         & 31.8         & 37.4         & 37.4         & 41.2         & 40.4         & 42.1         & 43.3         & 51.4         & 49.6      & 43.2   \\
CGDP+FSCN~\cite{li2021few}             & ResNet-101                  & 40.7         & 45.1         & 46.5         & 57.4         & 62.4         & 27.3         & 31.4         & 40.8         & 42.7         & 46.3         & 31.2         & 36.4         & 43.7         & 50.1         & 55.6      & 43.8   \\
CME~\cite{li2021beyond}                & ResNet-101                  & 41.5         & 47.5         & 50.4         & 58.2         & 60.9         & 27.2         & 30.2         & 41.4         & 42.5         & 46.8         & 34.3         & 39.6         & 45.1         & 48.3         & 51.5      & 44.4   \\
SRR-FSD~\cite{zhu2021semantic}         & ResNet-101                  & 47.8         & 50.5         & 51.3         & 55.2         & 56.8         & 32.5         & 35.3         & 39.1         & 40.8         & 43.8         & 40.1         & 41.5         & 44.3         & 46.9         & 46.4      & 44.8  \\
FSOD-UP~\cite{wu2021universal}         & ResNet-101                  & 43.8         & 47.8         & 50.3         & 55.4         & 61.7         & 31.2         & 30.5         & 41.2         & 42.2         & 48.3         & 35.5         & 39.7         & 43.9         & 50.6         & 53.5      & 45.0   \\
FSCE~\cite{sun2021fsce}                & ResNet-101                  & 44.2         & 43.8         & 51.4         & 61.9         & 63.4         & 27.3         & 29.5         & 43.5         & 44.2         & 50.2         & 37.2         & 41.9         & 47.5         & 54.6         & 58.5      & 46.6  \\
QA-FewDet~\cite{han2021query}          & ResNet-101                  & 42.4         & 51.9         & 55.7         & 62.6         & 63.4         & 25.9         & 37.8         & 46.6         & 48.9         & 51.1         & 35.2         & 42.9         & 47.8         & 54.8         & 53.5      & 48.0   \\
FADI~\cite{cao2021few}                 & ResNet-101                  & 50.3         & 54.8         & 54.2         & 59.3         & 63.2         & 30.6         & 35.0         & 40.3         & 42.8         & 48.0         & 45.7         & 49.7         & 49.1         & 55.0         & {\bf 59.6}& 49.2   \\
Zhang et al.~\cite{zhang2021accurate}  & ResNet-101                  & 48.6         & 51.1         & 52.0         & 53.7         & 54.3         & {\bf 41.6}   & \ul{45.4}    & 45.8         & 46.3         & 48.0         & 46.1         & \ul{51.7}    & 52.6         & 54.1         & 55.0      & 49.8   \\    
Meta FR-CNN~\cite{han2022meta}         & ResNet-101                  & 43.0         & 54.5         & 60.6         & \ul{66.1}    & \ul{65.4}    & 27.7         & 35.5         & 46.1         & 47.8         & \ul{51.4}    & 40.6         & 46.4         & \ul{53.4}    & {\bf 59.9}   & 58.6      & 50.5   \\
DeFRCN~\cite{qiao2021defrcn}           & ResNet-101                  & \ul{53.6}    & \ul{57.5}    & \ul{61.5}    & 64.1         & 60.8         & 30.1         & 38.1         & \ul{47.0}    & {\bf 53.3}   & 47.9         & \ul{48.4}    & 50.9         & 52.3         & 54.9         & 57.4      & \ul{51.9}  \\
{\bf VFA (Ours)}                       & ResNet-101                  & {\bf 57.7}   & {\bf 64.6}   & {\bf 64.7}   & {\bf 67.2}   & {\bf 67.4}   & \ul{41.4}    & {\bf 46.2}   & {\bf 51.1}   & \ul{51.8}    & {\bf 51.6}   & {\bf 48.9}   & {\bf 54.8}   & {\bf 56.6}   & \ul{59.0}    & \ul{58.9} & {\bf 56.1} \\
\bottomrule
\end{tabular}
}

    \caption{Results on PASCAL VOC. The results are sorted by the averaged score (Avg.). See our appendix for the generalized FSOD results.}
    \label{tab:pascal_voc_sota}
\end{table*}

\subsection{Variational Feature Aggregation}
\label{subsec:vfa}

Prior works usually encode support examples into single feature vectors that are difficult to represent the whole class distribution. 
Especially when the data is scarce and example's variations are large, we cannot make an accurate estimation of class centers. 
Inspired by recent progress in variational feature learning~\cite{lin2018deep,zhang2019variational,xu2021variational}, we transform support features into class distributions with VAEs. 
Since the estimated distribution is not biased to specific examples, features sampled from the distribution are robust to the variance of support examples. 
Then we can sample class-level features for robust feature aggregation.
The framework of VFA is shown in Fig.~\ref{fig:vfa}.

\noindent {\bf Variational Feature Learning.} Formally, we aim to transform the support feature $S$ into a class distribution $\mathcal{N}$, and sample the variational feature $z$ from $\mathcal{N}$ for feature aggregation. 
We optimize the model in a similar way to VAEs, but our goal is to sample the latent variable $z$ instead of the reconstructed feature $S^{'}$. Following the definition of VAEs, we assume $z$ is generated from a prior distribution $p(z)$ and $S$ is generated from a conditional distribution $p(S|z)$. As the process is hidden and $z$ is unknown, we model the posterior distribution with variational inference. More specifically, we approximate the true posterior distribution $p(z|S)$ with another distribution $q(z|S)$ by minimizing the Kullback-Leibler (KL) divergence:
\begin{align}
    D_{\textrm{KL}}(q(z|S)||p(z|S)) = \int q(z|S) \log \frac{q(z|S)}{p(z|S)},
\end{align}
which is equivalent to maximizing the evidence lower bound (ELBO):
\begin{equation} \label{eq:elbo}
\resizebox{.9\hsize}{!}{
    $ELBO=\mathbb{E}_{q(z|S)}[\log p(S|z))] - D_{\textrm{KL}}(q(z|S)||p(z)).$
}
\end{equation}

Here we assume the prior distribution of $z$ is a centered isotropic multivariate Gaussian, $p(z)=\mathcal{N}(0, I)$, and set the posterior distribution $q(z|S)$ to be a multivariate Gaussian with diagonal covariance: $q(z|S)=\mathcal{N}(\mu, \sigma)$. The parameters $\mu$ and $\sigma$ can be implemented by a feature encoder $\mathcal{F}_{enc}$: $\mu, \sigma=\mathcal{F}_{enc}(S)$. Then we obtain the variational feature $z$ with the reparameterization trick~\cite{kingma2013auto}: $z=\mu + \sigma \cdot \epsilon$, where $\epsilon \sim \mathcal{N}(0, I)$. The first term of Eq.~\ref{eq:elbo} can be simplified to a reconstruction loss $\mathcal{L}_{rec}$ which is usually defined as the L2 distance between the input $S$ and the reconstructed target $S^{'}$,
\begin{align}
    \mathcal{L}_{rec}=\|S-S^{'}\|=\|S-\mathcal{F}_{dec}(z)\|,
\end{align}
where $\mathcal{F}_{dec}$ denotes a feature decoder. As for the second term of Eq.~\ref{eq:elbo}, we directly minimize the KL divergence of $q(z|S)$ and $p(z)$,
\begin{align}
    \mathcal{L}_{\textrm{KL}}=D_{\textrm{KL}}(q(z|S)||p(z)),
\end{align}
which forces the variation feature $z$ to follow a normal distribution.

By optimizing the two objectives, $\mathcal{L}_{rec}$ and $\mathcal{L}_{\textrm{KL}}$, we transform the support feature $S$ into a distribution $\mathcal{N}$. Then we can sample the variational feature $z$ from $\mathcal{N}$. Since $z$ still lacks class-specific information, we apply a {\bf consistency loss} $\mathcal{L}_{cons}$ to the reconstructed feature $S^{'}$, which is defined as the cross-entropy between $S^{'}$ and its class label $c$,
\begin{align}
    \mathcal{L}_{cons} = \mathcal{L}_{\textrm{CE}}(\mathcal{F}_{cls}^{S^{'}}(S^{'}), c),
\end{align}
where $\mathcal{F}_{cls}^{S^{'}}$ denotes a linear classifier.
The introduction of $\mathcal{L}_{cons}$ transforms the learned distributions into class-specific distributions. The support feature $S_i$ is forced to approximate a parameterized distribution $\mathcal{N}(\mu_i, \sigma_i)$ of class $i$, so that the sampled $z$ can preserve class-specific information. 

\noindent {\bf Variational Feature Aggregation.} Since the support features are transformed into class distributions, we can sample features from the distribution and aggregate them with query features. Compared with the original support feature $S$ and reconstructed feature $S^{'}$, the latent variable $z$ contains more generic features of the class~\cite{zhang2019variational,lin2018deep}, which is robust to the variance of support examples.

Specifically, VFA follows the class-agnostic approach in CAA but aggregates the query feature $Q$ with a variational feature $z$. Given a query feature $Q_i$ of class $i$ and support feature $S_j$ of class $j$, we firstly approximate the class distribution $\mathcal{N}(\mu_j, \sigma_j)$ and sample a variational feature $z_j=\mu_j+\sigma_j$ from $\mathcal{N}(\mu_j, \sigma_j)$. Then we aggregate them together with the following equation:
\begin{align}\label{eq:vfa}
    \widetilde{Q}_{ij} = \mathcal{A}(Q_i, z_j) = Q_i \odot {\rm sig}(z_j),
\end{align}
where $\odot$ means channel-wise multiplication and sig is short for the \textit{sigmoid} operation.
In the training phase, we randomly select a support feature $S_j$ (\emph{i.e.}, one support class $j$) for aggregation.
In the testing phase (especially $K>1$), we average $K$ support features of class $j$ into one $\bar{S}_j$, and approximate the distribution $\mathcal{N}(\mu_j,\sigma_j)$ with the averaged feature, $\mu_j, \sigma_j = \mathcal{F}_{enc}(\bar{S}_j)$.
Instead of adopting complex distribution estimation methods, we find the averaging approach works well in our method.

\noindent {\bf Network and Objective.} VFA only introduces a light encoder $\mathcal{F}_{enc}$ and decoder $\mathcal{F}_{dec}$. $\mathcal{F}_{enc}$ contains a linear layer and two parallel linear layers to produce $\mu$ and $\sigma$, respectively. $\mathcal{F}_{dec}$ consists of two linear layers to generate the reconstructed feature $S^{'}$. We keep all layers the same dimension (2048 by default). VFA is trained in an end-to-end manner with the following multi-task loss:
\begin{align}
    \mathcal{L} = \mathcal{L}_{rpn} + \mathcal{L}_{reg} + \mathcal{L}_{cls} + \mathcal{L}_{cons} + \mathcal{L}_{rec} + \alpha \mathcal{L}_{KL},
\end{align}
where $\mathcal{L}_{rpn}$ is the total loss of RPN, $\mathcal{L}_{reg}$ is the regression loss, and $\alpha$ is a weight coefficient ($\alpha$=2.5$\times$10$^{-4}$ by default).

\subsection{Classification-Regression Decoupling}
\label{subsec:crd}

Generally, the detection head $\mathcal{F}_{D}$ contains a shared feature extractor $\mathcal{F}_{share}$ and two separate network $\mathcal{F}_{cls}$ and $\mathcal{F}_{reg}$ for classification and regression, respectively. 
In previous works, the aggregated feature is fed to $\mathcal{F}_{D}$ to produce both classification scores and bounding boxes. However, the classification task requires translation-invariant features, while regression needs translation-covariant features~\cite{qiao2021defrcn}. 
Since support features are always translation-invariant to represent class centers, the aggregated feature harms the regression task. 
Therefore, we decouple the two tasks in the detection head.
Let $Q$ and $\widetilde{Q}$ denote the original and aggregated query features. Previous methods take $\widetilde{Q}$ for both tasks, where the classification score $\mathbf{p}$ and predicted bounding boxes $\mathbf{b}$ are defined as:
\begin{align}
    \mathbf{p}=\mathcal{F}_{cls}(\mathcal{F}_{share}(\widetilde{Q})), \mathbf{b}=\mathcal{F}_{reg}(\mathcal{F}_{share}(\widetilde{Q})).
\end{align}
To decouple these tasks, we adopt separate feature extractors and use the original query feature $Q$ for regression,
\begin{align}
    \mathbf{p}=\mathcal{F}_{cls}(\mathcal{F}_{share}^{cls}(\widetilde{Q})), \mathbf{b}=\mathcal{F}_{reg}(\mathcal{F}_{share}^{reg}(Q)),
\end{align}
where $\mathcal{F}_{share}^{cls}$ and $\mathcal{F}_{share}^{reg}$ are the feature extractor for classification and regression, respectively.

\section{Experiments and Analysis}

\subsection{Experimental Setting}

\noindent {\bf Datasets.}
We evaluate our method on PASCAL VOC~\cite{everingham2010pascal} and COCO~\cite{lin2014microsoft}, following previous works~\cite{kang2019few,wang2020frustratingly}. We use the data splits and annotations provided by TFA~\cite{wang2020frustratingly} for a fair comparison.
For PASCAL VOC, we split 20 classes into three groups, where each group contains 15 base classes and 5 novel classes. For each novel set, we have $K$=\{1, 2, 3, 5, 10\} shots settings. 
For COCO, we set 60 categories disjoint with PASCAL VOC as base classes and the remaining 20  as novel classes. We have $K$=\{10, 30\} shots settings.

\noindent {\bf Evaluation Metrics.} For PASCAL VOC, we report the Average Precision at IoU=0.5 of base classes (bAP) and novel classes (nAP). For COCO, we report the mean AP at IoU=0.5:0.95 of novel classes (nAP).

\noindent {\bf Implementation Details.} We implement our method with Mmdetection~\cite{mmdetection}. The backbone is ResNet-101~\cite{he2016resnet} pre-trained on ImageNet~\cite{russakovsky2015imagenet}.
We adopt SGD as the optimizer with batch size 32, learning rate 0.02, momentum 0.9 and weight decay 1e-4. The learning rate is changed to 0.001 in the few-shot fine-tuning stage. We fine-tune the model with \{400, 800, 1200, 1600, 2000\} iterations for $K$=\{1, 2, 3, 5, 10\} shots in PASCAL VOC, and \{10000, 20000\} iterations for $K$=\{10, 30\} shots in COCO. We keep other hyper-parameters the same as Meta R-CNN~\cite{yan2019meta} if not specified.

\subsection{Main Results}

\begin{table}[t]
\begin{center}
    \small
    \begin{tabular}{l|cc}
        \toprule
        Method / Shots                         & 10            & 30              \\ \midrule
        \emph{Fine-tuning} &&\\          
        MPSR~\cite{wu2020multi}                & 9.8           & 14.1            \\
        TFA w/ cos~\cite{wang2020frustratingly}& 10.0          & 13.7            \\
        Retentive~\cite{fan2021generalized}    & 10.5          & 13.8            \\
        FSOD-UP~\cite{wu2021universal}         & 11.0          & 15.6             \\
        SRR-FSD~\cite{zhu2021semantic}         & 11.3          & 14.7             \\
        CGDP+FSCN~\cite{li2021few}             & 11.3          & 15.1             \\
        FSCE~\cite{sun2021fsce}                & 11.9          & 16.4             \\
        FADI~\cite{cao2021few}                 & 12.2          & 16.1             \\
        DeFRCN~\cite{qiao2021defrcn}           & 18.5          & 22.6            \\
        \midrule          
        \emph{Meta-learning} &&\\
        FSRW~\cite{kang2019few}                & 5.6           & 9.1              \\          
        MetaDet~\cite{wang2019meta}            & 7.1           & 11.3             \\
        Meta R-CNN~\cite{yan2019meta}          & 8.7           & 12.4               \\
        QA-FewDet~\cite{han2021query}          & 11.6          & 16.5             \\    
        FSDetView~\cite{xiao2020few}           & 12.5          & 14.7             \\
        Meta FR-CNN~\cite{han2022meta}         & 12.7          & 16.6             \\
        DCNet~\cite{hu2021dense}               & 12.8          & 18.6             \\
        CME~\cite{li2021beyond}                & 15.1          & 16.9             \\
        {\bf VFA (Ours)}                       & {\bf 16.2}    & {\bf 18.9}       \\
        \bottomrule
        \end{tabular}
    \end{center}

    \caption{Results on COCO. The backbone is the same as Tab.~\ref{tab:pascal_voc_sota}. The results are sorted by 10-shot nAP. See our appendix for the generalized FSOD results.}
    \label{tab:coco_sota}
\end{table}

\noindent {\bf PASCAL VOC.} As shown in Tab.~\ref{tab:pascal_voc_sota}, VFA significantly outperforms existing methods. VFA achieves the best (13/16) or second-best (3/16) results on all settings. In Novel Set 1, VFA outperforms previous best results by 3.2\%$\sim$7.1\%. Our 2-shot result even surpasses previous best 10-shot results (64.6\% \emph{vs.} 63.4\%), which indicates that our method is more robust to the variance of few-shot examples.
Besides, we notice that our gains are stable and consistent. This phenomenon demonstrates that VFA is not biased to specified class sets and can be generalized to more common scenarios. Furthermore, VFA obtains a 56.1\% average score and surpasses the second-best result by 4.2\%, which further demonstrates its effectiveness.

\noindent {\bf COCO.} As shown in Tab.~\ref{tab:coco_sota}, VFA achieves the best nAP among meta-learning based methods and second-best results among all methods. 
We notice that a fine-tuning based method, DeFRCN~\cite{qiao2021defrcn}, outperforms our method in nAP. To concentrate on the feature aggregation module in meta-learning, we do not utilize advanced techniques, \emph{e.g.}, the gradient decoupled layer~\cite{qiao2021defrcn} in DeFRCN. We believe the performance of VFA can be further boosted with more advanced techniques.

\begin{table}[t]
    \begin{center}
    \small
    \begin{tabular}{lccc|ccc}
    \toprule
    \multirow{2}{*}{Method} & \multirow{2}{*}{CRD}      & \multirow{2}{*}{CAA}      & \multirow{2}{*}{VFA}      & \multicolumn{3}{c}{Shots}  \\
                            &                           &                           &                           & 1        & 3        & 5          \\ \midrule 
    \multicolumn{4}{l|}{Meta R-CNN++}                                                                           & 42.0     & 56.5     & 58.3       \\ \midrule
    \multirow{3}{*}{Ours}   & \checkmark                &                           &                           & 46.0     & 61.7     & 62.3       \\
                            & \checkmark                & \checkmark                &                           & 51.3     & 62.8     & 66.4       \\
                            & \checkmark                & \checkmark                & \checkmark                &{\bf 57.7}&{\bf 64.7}&{\bf 67.2}  \\ \bottomrule
    \end{tabular}
\end{center}

    \caption{Effect of different modules.}
    \label{tab:ablation_modules}
\end{table}

\begin{figure}
    \centering
    \includegraphics[width=\linewidth]{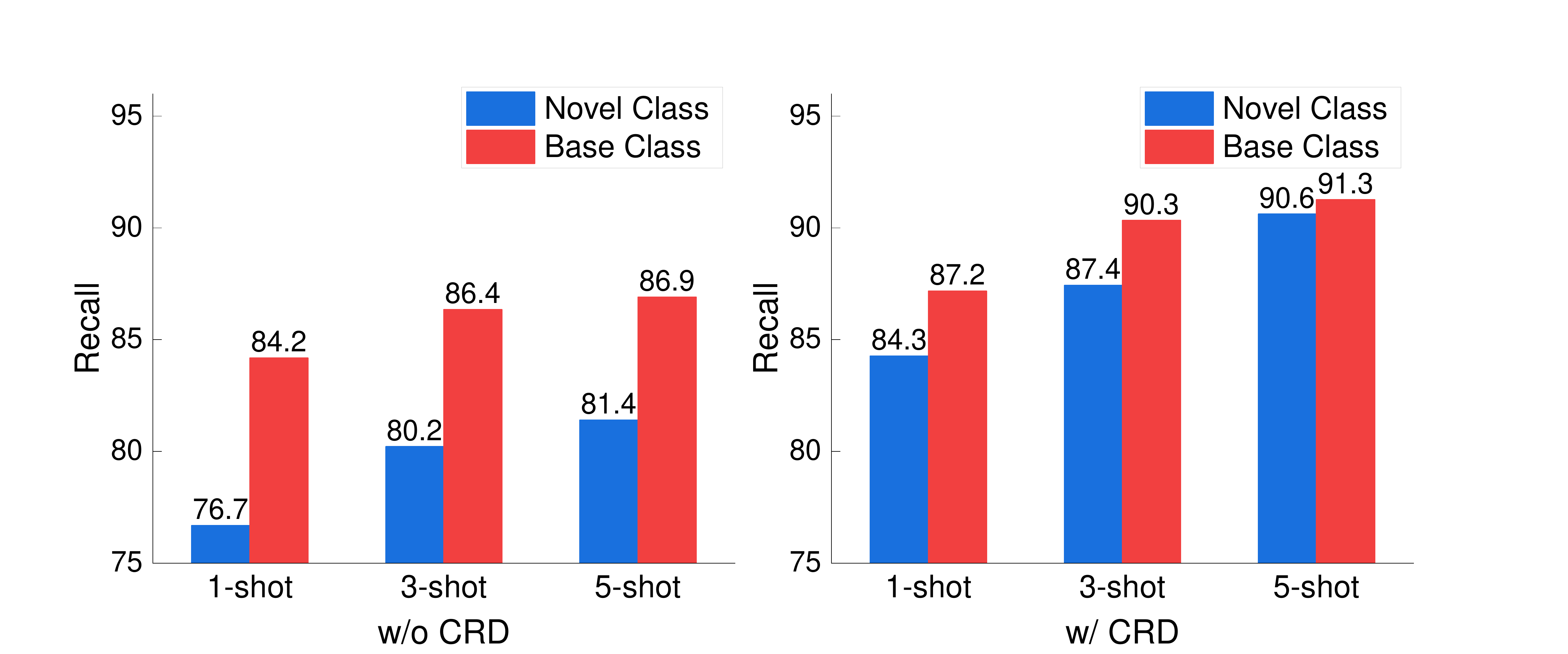} 
    \caption{Comparisons of recall without/with CRD.}
    \label{fig:crd}   
\end{figure}

\begin{figure}[t]
    \centering
    \includegraphics[width=\linewidth]{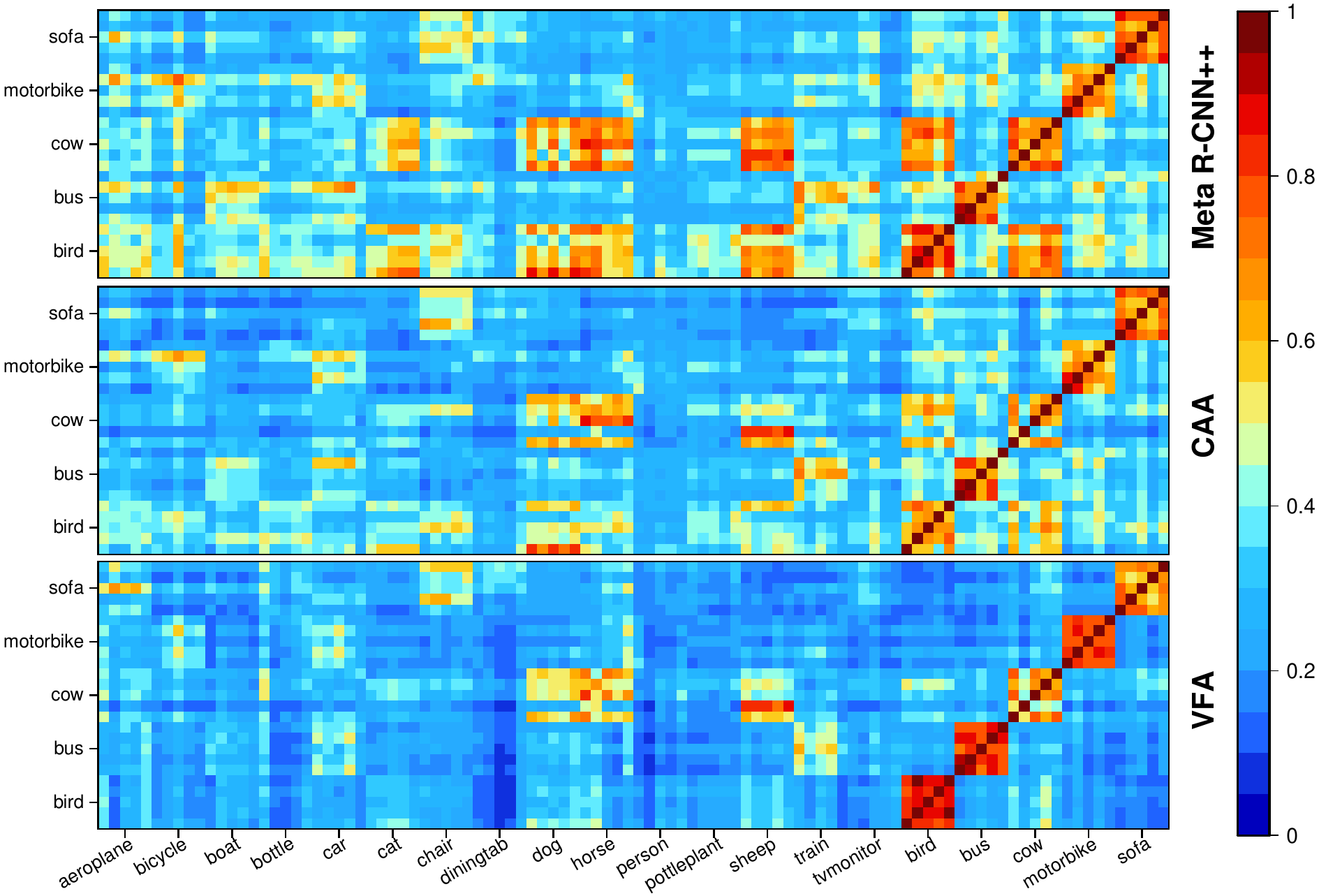}
    \caption{Similarity matrix visualization. We calculate cosine similarities of support features in the 5-shot setting of PASCAL VOC Novel Set 1. \textit{sofa, motorbike, cow, bus} and \textit{bird} are novel classes. Warmer color denotes higher similarity. Zoom in for details.}
    \label{fig:confusion_matrix}
\end{figure}

\begin{figure}[t]
    \centering
    \includegraphics[width=\linewidth]{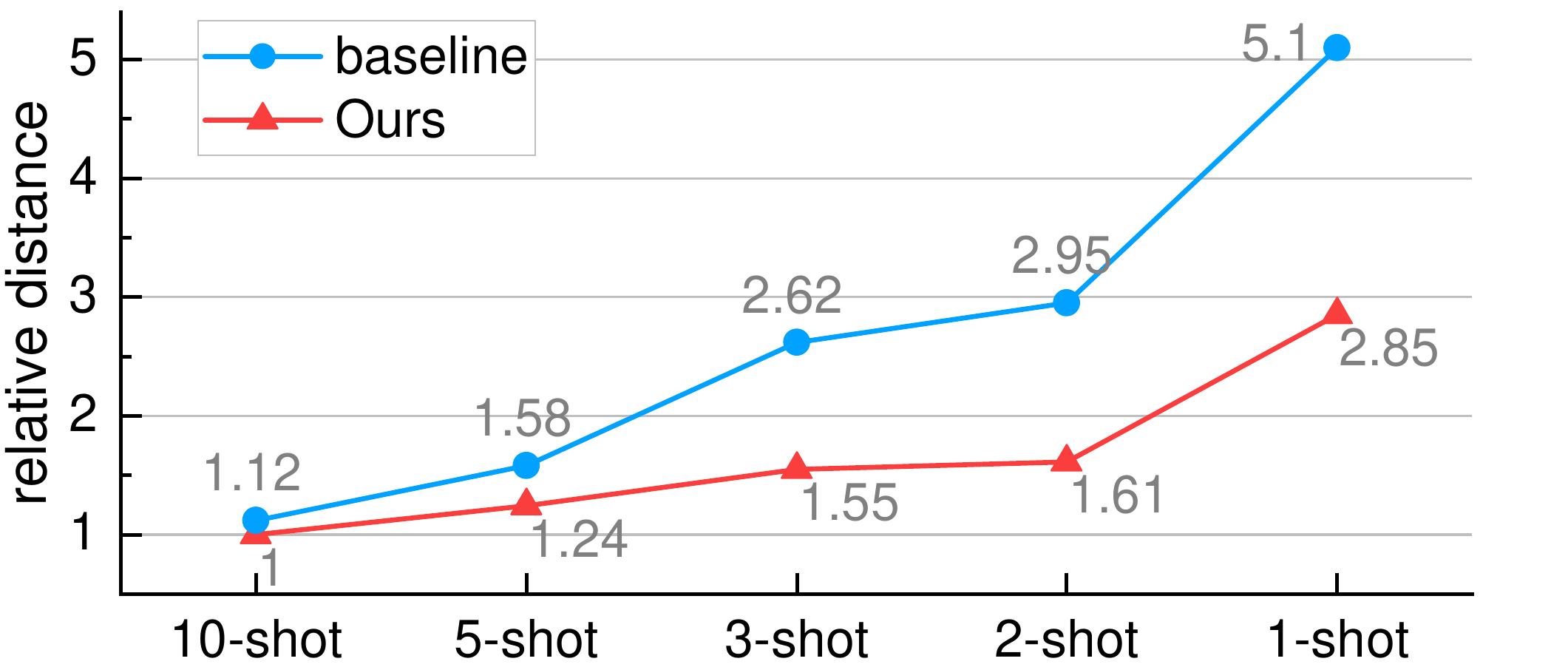}
    \caption{The distance from the estimated prototype of $K$-shot examples to the real class center. For each novel class, we take the mean feature of all training examples as its real class center. Our 10-shot result is the reference distance, while other results are relative distances. We only report the averaged distance of all novel classes for simplicity.}
    \label{fig:vfa_center_acc}
\end{figure}

\begin{table}[t]
\begin{center}
\small
\begin{tabular}{lc|cccccc}
\toprule
\multicolumn{2}{l|}{Features} & $S$ & $S^{'}$ & $\mu$ & $\sigma$ & $\widetilde{z}$ & $z$ \\ \midrule
\multicolumn{2}{l|}{bAP}     & {\bf 78.8}  & 78.1   & 78.6  & 78.3  & 78.0     & 78.6    \\
\midrule
\multirow{4}{*}{nAP}   & 1   & 55.2  & 54.4   & 56.6  & 55.4  & 53.0     & {\bf 57.7}    \\
                       & 3   & 63.7  & 63.6   & 63.7  & {\bf 64.9}  & 63.2     & 64.7    \\
                       & 5   & 66.6  & 66.9   & 66.7  & 66.9  & 66.3     & {\bf 67.2}    \\
                       & avg.& 61.8  & 61.6   & 62.3  & 62.4  & 60.8     & {\bf 63.2}    \\
\bottomrule
\end{tabular}
\end{center}
    \vspace{-2mm}
    \caption{Comparisons of different support features. $S$ and $S^{'}$ are the original and reconstructed features. $\mu$, $\sigma$, $\widetilde{z}=\mu + \epsilon \cdot \sigma$ and $z=\mu + \sigma$ are latent variables. avg.: The average score.}
    \label{tab:feature_aggregation}
\end{table}

\begin{table}[t]
    \centering
\small
\begin{tabular}{cc|ccc}
\toprule
\multicolumn{2}{l|}{Setting / Shots}                                       & 1           & 3           & 5           \\ \midrule
w/o VFA &                                   & 51.3        & 62.8        & 66.4        \\ \midrule
\multirow{3}{*}{w/ VFA} & \multicolumn{1}{l|}{w/o $\mathcal{L}_{cons}$}    & 53.6        & 64.3        & 66.7        \\
                        & \multicolumn{1}{l|}{$\mathcal{L}_{cons}$ on $S$} & 52.9        & 64.1        & {\bf 67.3}  \\
                        & $\mathcal{L}_{cons}$ on $S^{'}$                  & {\bf 57.7}  & {\bf 64.7}  & 67.2        \\ \bottomrule

\end{tabular}
   \caption{Effect of $\mathcal{L}_{cons}$. w/o: without. $\mathcal{L}_{cons}$ on $S$/$S^{'}$: apply $\mathcal{L}_{cons}$ to $S$ or $S^{'}$. The results are averages of multiple runs.}
    \label{tab:consistency_loss}
\end{table}

\subsection{Ablation Studies}

We conduct a series of ablation experiments on Novel Set 1 of PASCAL VOC.

\noindent {\bf Effect of different modules.}
As shown in Tab.~\ref{tab:ablation_modules}, we evaluate the effect of different modules by gradually applying the proposed modules to Meta R-CNN++. Although Meta R-CNN++ is competitive enough, we show {\bf CRD} improves the performance on nAP, where the absolute gains exceed 4\%. Besides, we find CRD significantly improves the recall on all classes (Fig.~\ref{fig:crd}) and narrows the gap between base and novel classes because it uses separate networks to learn translation-invariant and -covariant features.
Then, we apply {\bf CAA} to the model and obtain further improvements. The confusions between different classes are reduced.
Finally, we build {\bf VFA} and achieve a new state-of-the-art. The 1-shot performance is even comparable with 5-shot Meta R-CNN++ in nAP, indicating that VFA is robust to the variance of support examples especially when the data is scarce.


\noindent {\bf Visual analysis of different feature aggregation.}
Fig.~\ref{fig:confusion_matrix} gives a visual analysis of different feature aggregation methods.
Due to the imbalance between base and novel classes, some novel classes are confused with base classes in Meta R-CNN++ (with CSA), \emph{e.g.}, a novel classe, \textit{cow} have higher similarity ($>$0.8) with \textit{horse} and \textit{sheep}. 
In contrast, CAA reduces class bias and confusion by learning class-agnostic representations. The inter-class similarities are also reduced so that a novel example will not be classified to base classes.
Finally, we use VFA to transforms support examples into class distributions.
By learning intra-class variances from abundant base examples, we can estimate novel classes' distributions even with a few examples.
In Fig.~\ref{fig:confusion_matrix} (bottom), we can see VFA significantly improves intra-class similarities. 

\noindent {\bf Robust and accurate class prototypes.} In the testing phase, detectors take the mean feature of $K$-shot examples as the class prototype. 
As shown in Fig.~\ref{fig:vfa_center_acc}, our estimated class prototypes are more robust and accurate than the baseline. The distances to real class centers do not increase much as the shot decreases, because our method can fully leverage base classes' distributions to estimate novel classes' distributions.
The prototypes sampled from distributions are robust to the variance of support examples.
While the baseline is sensitive to the number of support examples.

\noindent {\bf Which feature to aggregate?}
In Tab.~\ref{tab:feature_aggregation}, we explore different features for aggregation. All types of features achieve comparable performance on base classes but vary on novel classes. The performance of original feature $S$ and reconstructed feature $S^{'}$ lag behind the latent encoding $\mu$, $\sigma$ and $z$. We hypothesize that the latent encoding contains more class-generic features. Besides, $\widetilde{z}=\mu + \epsilon \cdot \sigma$ performs worst among these features due to its indeterminate inference process. Instead, a simplified version $z=\mu + \sigma$ achieves satisfactory results, which is the default setting of VFA.


\noindent {\bf Effect of $\mathcal{L}_{cons}$.} We use a shared VAE to encode support features but still need to preserve class-specific information. Therefore, we add a consistency loss $\mathcal{L}_{cons}$ to produce class-wise distributions.
Tab.~\ref{tab:consistency_loss} shows that $\mathcal{L}_{cons}$ is important for VFA. $\mathcal{L}_{cons}$ applied to $S^{'}$ forces the model to produce class-conditional distributions so that the latent variable $z$ can retrain meaningful information to represent class centers.

\noindent {\bf Design of VFA.} The variational feature encoder $\mathcal{F}_{enc}$ and decoder $\mathcal{F}_{dec}$ are not sensitive to the number and dimension of hidden layers. Please see our appendix for details.


\section{Conclusion}
This paper revisits feature aggregation schemes in meta-learning based FSOD and proposes Class-Agnostic Aggregation (CAA) and Variational Feature Aggregation (VFA). 
CAA can reduce class bias and confusion between base and novel classes; VFA transforms instance-wise support features into class distributions for robust feature aggregation. 
Extensive experiments on PASCAL VOC and COCO demonstrate our effectiveness.

\section{Acknowledgement}
This work was partially supported by National Nature Science Foundation of China under the grants No.U22B2011, No.41820104006, and No.61922065.

\bibliography{aaai23}

\newpage

\appendix


\section{Additional Main Results.}

\noindent {\bf Results on Generalized FSOD.}
We evaluate our method on the Generalized FSOD benchmark~\cite{wang2020frustratingly}. The result is an average of multiple random seeds. Following~\cite{qiao2021defrcn}, we report nAP of different methods with 10 random seeds. Since many methods only report their results on the traditional FSOD benchmarks, we collect as many methods that report the G-FSOD results as possible.
{\bf PASCAL VOC:} Similar to the results of our main paper, our method performs well on PASCAL VOC. As shown in Tab.~\ref{tab:gfsod_pascal_voc_sota}, our method achieves the best (12/15) or second-best (3/15) among all settings. Especially when the shot is low, our method shows significant improvements. For example, our 1-shot gains are 7.2\%, 4.2\% and 8.8\% on the Novel Set 1, 2 and 3, respectively.
{\bf COCO:} We also compare VFA with other methods on COCO, where our method achieves the second-best results on nAP. We notice that the gap between VFA and DeFRCN is narrowed in the G-FSOD setting (0.9\% \emph{vs.} 2.3\% on 10-shot nAP).

\begin{table*}[!t]
\setlength{\tabcolsep}{3.7pt}    
\small
\centering
\begin{tabular}{@{}l|ccccc|ccccc|ccccc|c@{}}
    \toprule
    \multirow{2}{*}{Method / Shots}        & \multicolumn{5}{c|}{Novel Set 1}                                         & \multicolumn{5}{c|}{Novel Set 2}                                         & \multicolumn{5}{c|}{Novel Set 3}                                       & \multirow{2}{*}{Avg.}    \\ 
                                           & 1            & 2            & 3            & 5            & 10           & 1            & 2            & 3            & 5            & 10           & 1            & 2            & 3            & 5            & 10        &    \\ \midrule
    FRCN-ft~\cite{ren2017faster}           & 9.9          & 15.6         & 21.6         & 28.0         & 52.0         & 9.4          & 13.8          & 17.4        & 21.9         & 39.7         & 8.1          & 13.9         & 19.0         & 23.9         & 44.6      &  22.6      \\
    FSRW~\cite{kang2019few}                & 14.2         & 23.6         & 29.8         & 36.5         & 35.6         & 12.3         & 19.6          & 25.1        & 31.4         & 29.8         & 12.5         & 21.3         & 26.8         & 33.8         & 31.0      &  25.6  \\
    TFA w/ cos~\cite{wang2020frustratingly}& 25.3         & 36.4         & 42.1         & 47.9         & 52.8         & 18.3         & 27.5          & 30.9        & 34.1         & 39.5         & 17.9         & 27.2         & 34.3         & 40.8         & 45.6      &  34.7  \\
    FSDetView~\cite{xiao2020few}           & 24.2         & 35.3         & 42.2         & 49.1         & 57.4         & 21.6         & 24.6          & 31.9        & 37.0         & 45.7         & 21.2         & 30.0         & 37.2         & 43.8         & 49.6      &  36.7  \\
    DCNet~\cite{hu2021dense}               & 33.9         & 37.4         & 43.7         & 51.1         & 59.6         & 23.2         & 24.8          & 30.6        & 36.7         & 46.6         & 32.3         & 34.9         & 39.7         & 42.6         & 50.7      &  39.2  \\
    FSCE~\cite{sun2021fsce}                & 32.9         & 44.0         & 46.8         & 52.9         & 59.7         & 23.7         & 30.6         & 38.4         & 43.0         & 48.5         & 22.6         & 33.4         & 39.5         & 47.3         & 54.0      &  41.2   \\
    DeFRCN~\cite{qiao2021defrcn}           & \underline{40.2}  & \underline{53.6}  & \underline{58.2}  & \underline{63.6}  & \textbf{66.5}   & \underline{29.5}  & \textbf{39.7}    & \underline{43.4} & \underline{48.1}  & \textbf{52.8}   & \underline{35.0}  & \underline{38.3}  & \underline{52.9}  & \underline{57.7}  & \textbf{60.8}&  \underline{49.4}  \\
    {\bf VFA (Ours)}                       & \textbf{47.4}   & \textbf{54.4}   & \textbf{58.5}   & \textbf{64.5}   & \textbf{66.5}   & \textbf{33.7}   & \underline{38.2}   & \textbf{43.5}  & \textbf{48.3}   & \underline{52.4}  & \textbf{43.8}   & \textbf{48.9}   & \textbf{53.3}   & \textbf{58.1}   & \underline{60.0}&  \textbf{51.4}  \\
    \bottomrule
    \end{tabular}

    \caption{G-FSOD results on PASCAL VOC. The results are sorted by the averaged score (Avg.).}
    \label{tab:gfsod_pascal_voc_sota}
\end{table*}

\begin{table}[t!]
    \centering
\begin{center}
\begin{tabular}{l|cc}
    \toprule
    \multirow{2}{*}{Method}                & \multicolumn{2}{c}{Shots}       \\
                                           & 10            & 30             \\ \midrule
    TFA w/ cos~\cite{wang2020frustratingly}& 9.1           & 12.1           \\
    FSDetView~\cite{xiao2020few}           & 10.7          & 15.9           \\
    FSCE~\cite{sun2021fsce}                & 11.1          & 15.3           \\
    DeFRCN~\cite{qiao2021defrcn}           & \textbf{16.8}    & \textbf{21.2}        \\
    {\bf VFA (Ours)}                       & \underline{15.9}   & \underline{18.4}   \\
    \bottomrule
    \end{tabular}
\end{center}
    \caption{G-FSOD results on COCO. The results are sorted by 10-shot nAP.}
    \label{tab:gfsod_coco_sota}
\end{table}


\section{Additional Ablation Studies}

\begin{figure}[t]
    \centering
    \includegraphics[width=\linewidth]{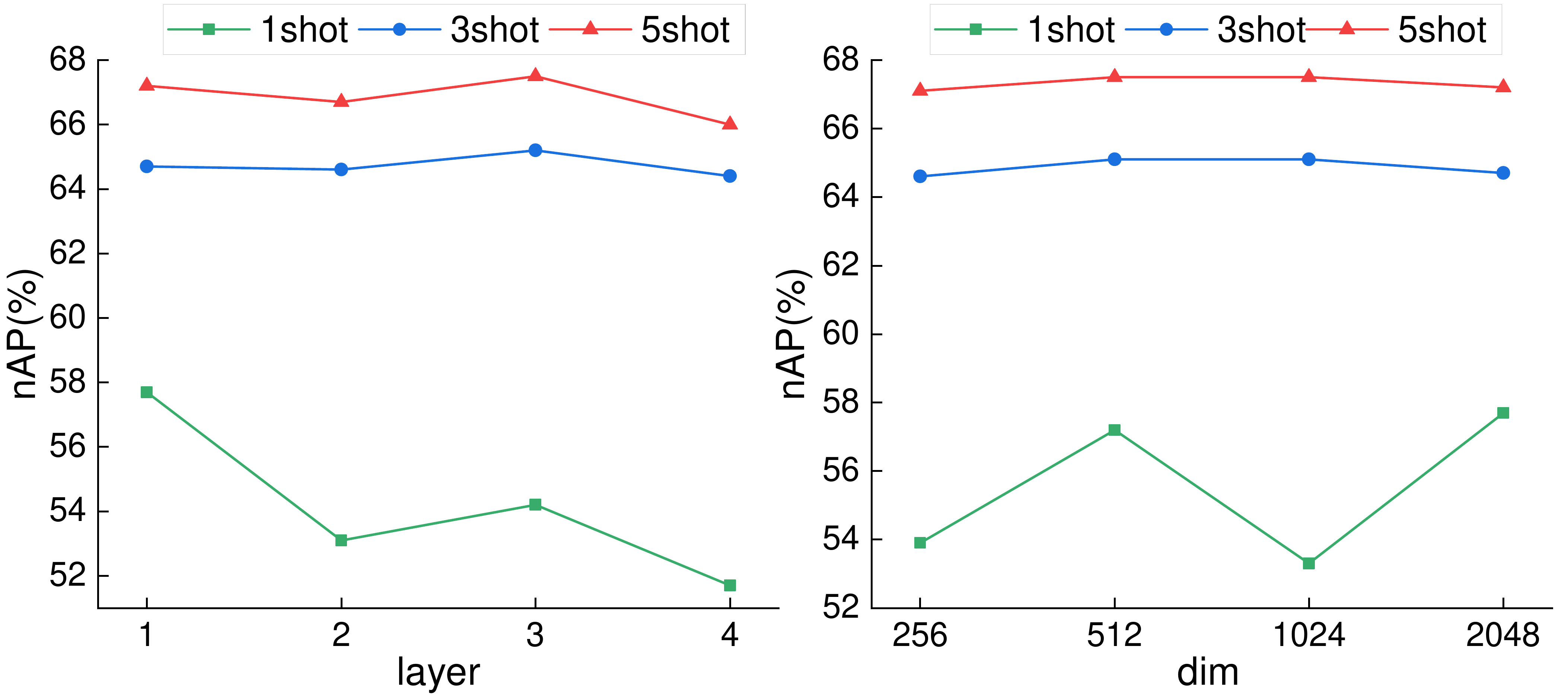} 
    \caption{Design of VFA. We explore different designs of $\mathcal{F}_{enc}$ and $\mathcal{F}_{dec}$. layer: the number of hidden layer. dim: the number of hidden channels.}
    \label{fig:vae_design}   
\end{figure}

\begin{table}[t!]
    \centering
\begin{tabular}{l|ccc|ccc}
\toprule
\multirow{2}{*}{$\mathcal{F}_{enc}$, $\mathcal{F}_{dec}$} & \multicolumn{3}{c|}{nAP} & \multicolumn{3}{c}{bAP} \\
                                   & 1      & 3      & 5     & 1      & 3      & 5     \\ \midrule
freeze                             & 57.6   & 64.5   & 67.1  & 71.5   & 75.9   & 76.7  \\
trainable                          & 57.7   & 64.7   & 67.2  & 71.6   & 76.0   & 76.7   \\
$\Delta$                           & -0.1   & -0.2   & -0.1  & -0.1   & -0.1   & 0.0    \\
\bottomrule
\end{tabular}
    \caption{Freeze or fine-tune $\mathcal{F}_{enc}$ and $\mathcal{F}_{dec}$ in VFA. $\Delta$: The difference between the first and second row.}
    \label{tab:vfa_freeze}
\end{table}

\noindent {\bf Design of VFA.}
By default, the feature encoder $\mathcal{F}_{enc}$ and decoder $\mathcal{F}_{dec}$ consist of one input layer and output layer of 1024-$d$. 
In Fig.~\ref{fig:vae_design}, we ablate the number of input layers and hidden channels. VFA is sensitive to these hyper-parameters when the shot is low (up to 4\% in 1 shot). The performance becomes more stable as the shot increases, e.g., the gap between different settings is reduced to 1\% in 3 and 5 shots.

\noindent {\bf VFA with/without fine-tuning.} In the few-shot fine-tuning stage, we fine-tune the variational feature encoder $\mathcal{F}_{enc}$ and decoder $\mathcal{F}_{dec}$ by default. Tab.~\ref{tab:vfa_freeze} shows that $\mathcal{F}_{enc}$ and $\mathcal{F}_{dec}$ can work without fine-tuning. The gap between two settings, freeze parameters \emph{vs.} trainable, is relatively small (about 0.1\%). The results indicate that the representation learned from base classes can be directly transferred to novel classes even without fine-tuning. 

\begin{figure*}[htb!]
    \centering
    \includegraphics[width=.9\textwidth]{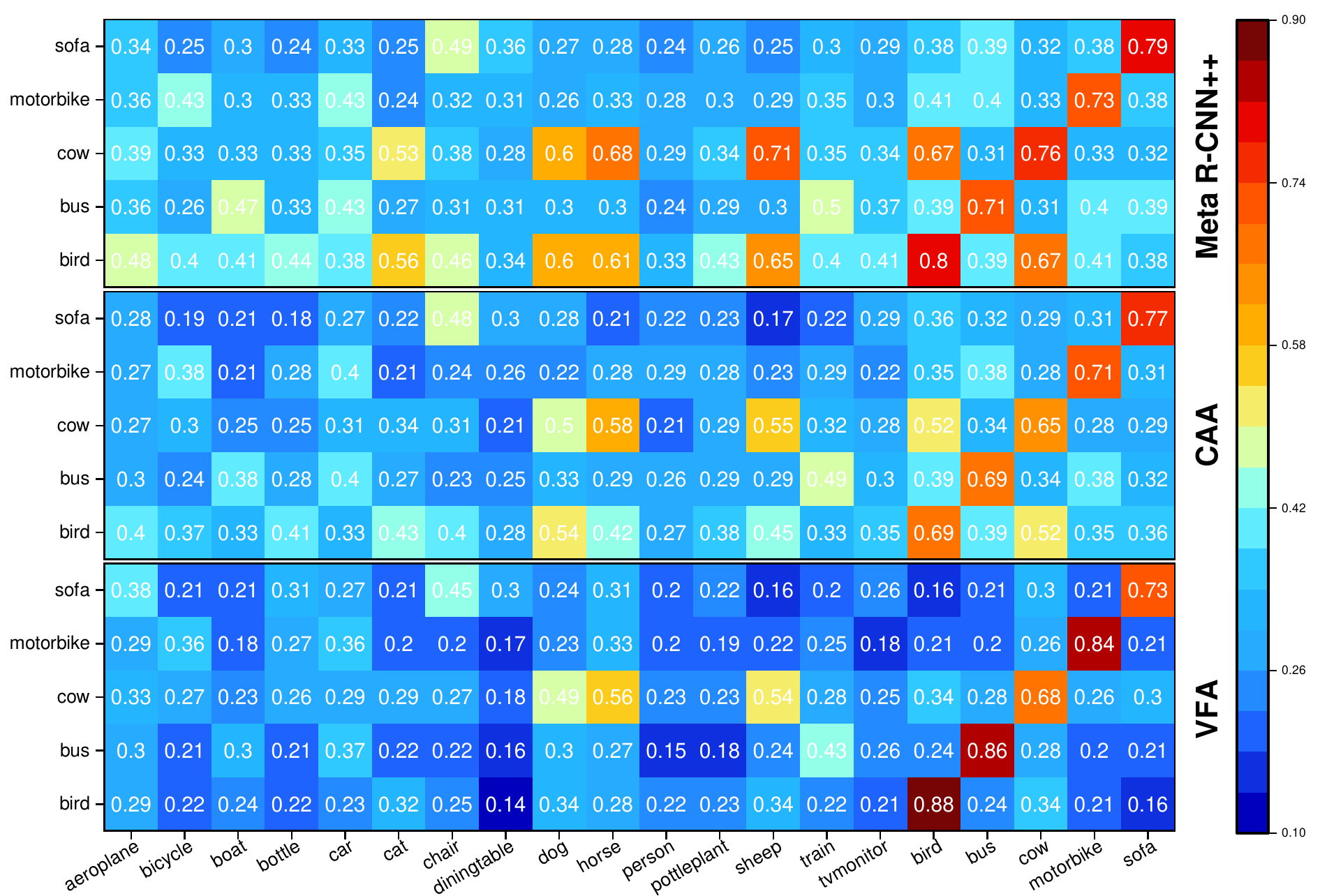}
    \caption{Analysis of different feature aggregation. We follow the same setting as Fig.5 of our main paper. Differently, the number in each cell is the averaged cosine similarity of 5 shot examples.}
    \label{fig:confusion_table}
\end{figure*}

\noindent {\bf More analysis of different feature aggregation.}
In the main paper, we give a visual analysis of different feature aggregation methods, i.e., CSA, CAA and VFA. 
Here we give a quantitative analysis of these methods, shown in Fig~\ref{fig:confusion_table}. Compared with CSA, CAA reduces class confusion between base and novel classes. For example, The similarity between \texttt{cow} and \texttt{sheep} is 0.71, near the intra-class similarity of \texttt{cow} (0.76). 
While in CAA, the similarity between \texttt{cow} and \texttt{sheep} is reduced to 0.55 and the gap of intra-class and inter-class similarity is enlarged to 0.10 (0.65 \emph{vs.} 0.55). By applying VFA to the model, the inter-class similarity is further reduced. For each novel class, the gap between intra-class and inter-class similarity is enlarged to 0.12$\sim$0.54 (the range is 0.05$\sim$0.3 in CSA). 
The results further demonstrate that our proposed CAA and VFA learn more discriminative and transferable features.

\section{Visualization}
\begin{figure*}[htb!]
    \centering
    \includegraphics[width=.9\textwidth]{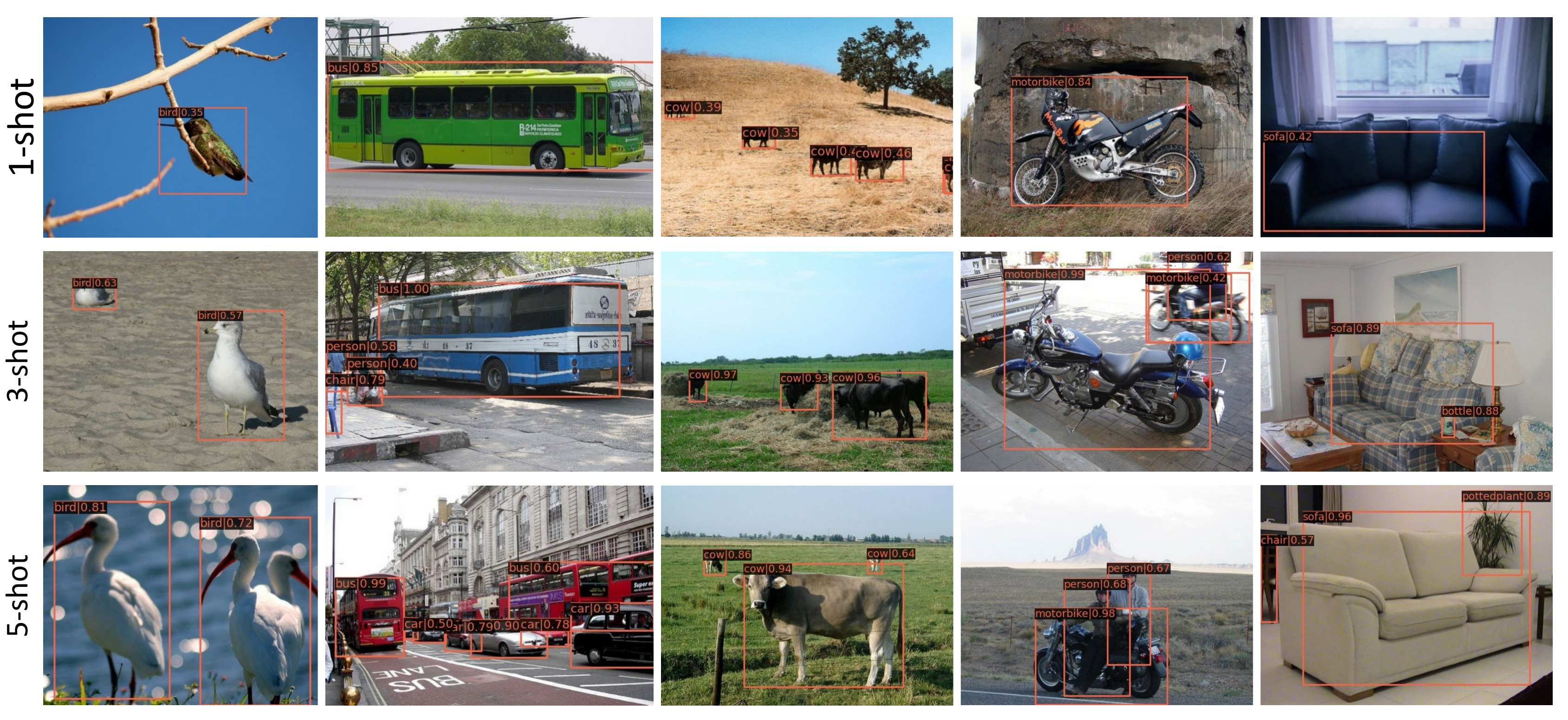}
    \caption{Visualization of some detection results. The model is trained on the \{1, 3, 5\} shots of PASCAL VOC Novel Set 1 and tested on the VOC07 \texttt{test} set.}
    \label{fig:visualization}
\end{figure*}

We visualize the detection results in Fig.~\ref{fig:visualization}. In the base training stage, we pre-train the model on base classes of PASCAL VOC. Then we fine-tune the model on the \{1, 3, 5\} shots of Novel Set 1 and visualize the detection results. As the support set grows, our model produces more confident results, \emph{e.g.}, the scores of detected novel objects are increasing from 1-shot to 5-shot.

\section{More Training Details}
Our method follows the two-stage training strategy in FSOD~\cite{yan2019meta,kang2019few}, \emph{i.e.}, base training and few-shot fine-tuning. In the {\bf base training} stage, we build a query dataset and a support dataset, where the query dataset contains the whole data from base classes and the support dataset is obtained by balanced sampling from the query dataset.
We train the model on the two datasets and update all network parameters.
In the {\bf few-shot fine-tuning} stage, the support dataset is usually the same as the query dataset with only $K$ shot instances. We only train (a) our $\mathcal{F}_{enc}$ and $\mathcal{F}_{dec}$ in VFA and (b) the last classification and regression layers. We freeze other parameters except for the Region Proposal Network (RPN) by default. RPN is fixed in our PASCAL VOC experiments but not frozen in COCO experiments because the model on COCO will not overfit to novel classes (10$\sim$30 shots).

\end{document}